\documentclass[10pt,conference]{IEEEtran}

\usepackage{microtype}
\usepackage{url}
\usepackage{amsmath}
\usepackage{amssymb}
\usepackage{array}
\usepackage{booktabs}
\usepackage{enumitem}
\usepackage{graphicx}
\usepackage{tabularx}
\usepackage{xcolor}
\usepackage{xspace}
\usepackage[T1]{fontenc}
\usepackage{textcomp}
\usepackage[numbers,sort&compress]{natbib}
\usepackage{hyperref}
\usepackage[most]{tcolorbox}
\usepackage{balance}

\definecolor{darkblue}{rgb}{0, 0, 0.5}
\hypersetup{colorlinks=true, citecolor=darkblue, linkcolor=darkblue, urlcolor=darkblue}

\title{Skill Coverage: A Test Adequacy Metric for Agent Skills}

\author{
\IEEEauthorblockN{Boyin Tan\textsuperscript{1}, Xiaowei Huang\textsuperscript{2}, Youcheng Sun\textsuperscript{1}}
\IEEEauthorblockA{\textsuperscript{1}\textit{Mohamed bin Zayed University of Artificial Intelligence}\\
Abu Dhabi, United Arab Emirates\\
\{Boyin.Tan,Youcheng.Sun\}@mbzuai.ac.ae}
\IEEEauthorblockA{\textsuperscript{2}\textit{University of Liverpool}\\
Liverpool, United Kingdom\\
Xiaowei.Huang@liverpool.ac.uk}
}

\begin{document}

\maketitle

\begin{abstract}
Agent skills encode reusable procedural knowledge for large language model
(LLM) agents, and existing benchmarks show that such skills can improve
task-level performance. However, a task outcome does not reveal which parts of a
reusable skill were exercised, nor whether the agent followed the relevant skill
instructions when those parts were exercised. This gap makes it unclear whether
a skill has been adequately tested, or whether observed task failures provide
actionable evidence for improving agent skill effectiveness.

To fill this gap, we introduce \textit{skill coverage}, a trajectory-based
test-adequacy metric for reusable agent skills. Our framework extracts skill
behavior constraints from each skill, translating natural-language skill
instructions into semi-structured constraints that specify the expected agent
behavior under particular conditions. It then determines whether each
constraint is covered by an agent trajectory and, for covered constraints,
assigns a \textit{Pass} or \textit{Fail} verdict according to the agent
behavior. We apply this framework to SkillsBench. The results show that agent
trajectories on the benchmark leaderboard cover only 38.66--45.51\% of the
extracted skill behavior constraints on average. We then use \textit{Fail}
verdicts to strengthen the corresponding skill content only by emphasizing the
original instructions that the agent failed to follow, and run the same tasks
with the strengthened skills. This emphasis yields an average 16.0\% recovery
rate of the failed tasks across the five agent--model rows. These results show
that skill coverage is both a test-adequacy metric and a fine-grained signal for
observing skill-use behavior. In failed tasks, failed constraint labels provide
actionable evidence for improving agent skill effectiveness. The project
website is available at \url{https://shuaijiumei.github.io/skillcoverage/}.
\end{abstract}


\newcommand{\sbc}{\textsc{SBC}\xspace}
\newcommand{\sbcs}{\textsc{SBC}s\xspace}

\section{Introduction}

Agent skills are structured instructions that give LLM agents procedural knowledge at inference time~\citep{anthropic2026agentskills,Li2026SkillsBenchBH}.
Skills encode standard operating procedures, domain conventions, and task heuristics as modular artifacts mediated by the agent harness~\citep{Sumers2023CognitiveAF, Sutton1999BetweenMA}.
Benchmarks show that skills can improve task performance across diverse domains~\citep{Li2026SkillsBenchBH,Zhong2026SkillLearnBenchBC,Liu2026HowWD}, while studies of public skill ecosystems show that skills are becoming reusable artifacts that can be shared across agents and tasks~\citep{Huang2025CASCADECA,Ying2026OpenSkillEvalAA}.
On ClawHub~\citep{clawhub2026topskills}, some well-known skills are downloaded thousands of times, such as \texttt{gog} and \texttt{nano-pdf} have received 187k and 114k downloads, respectively~\citep{clawhub2026gog,clawhub2026nanopdf}.

\begin{figure}[t]
    \centering
    \includegraphics[width=\columnwidth]{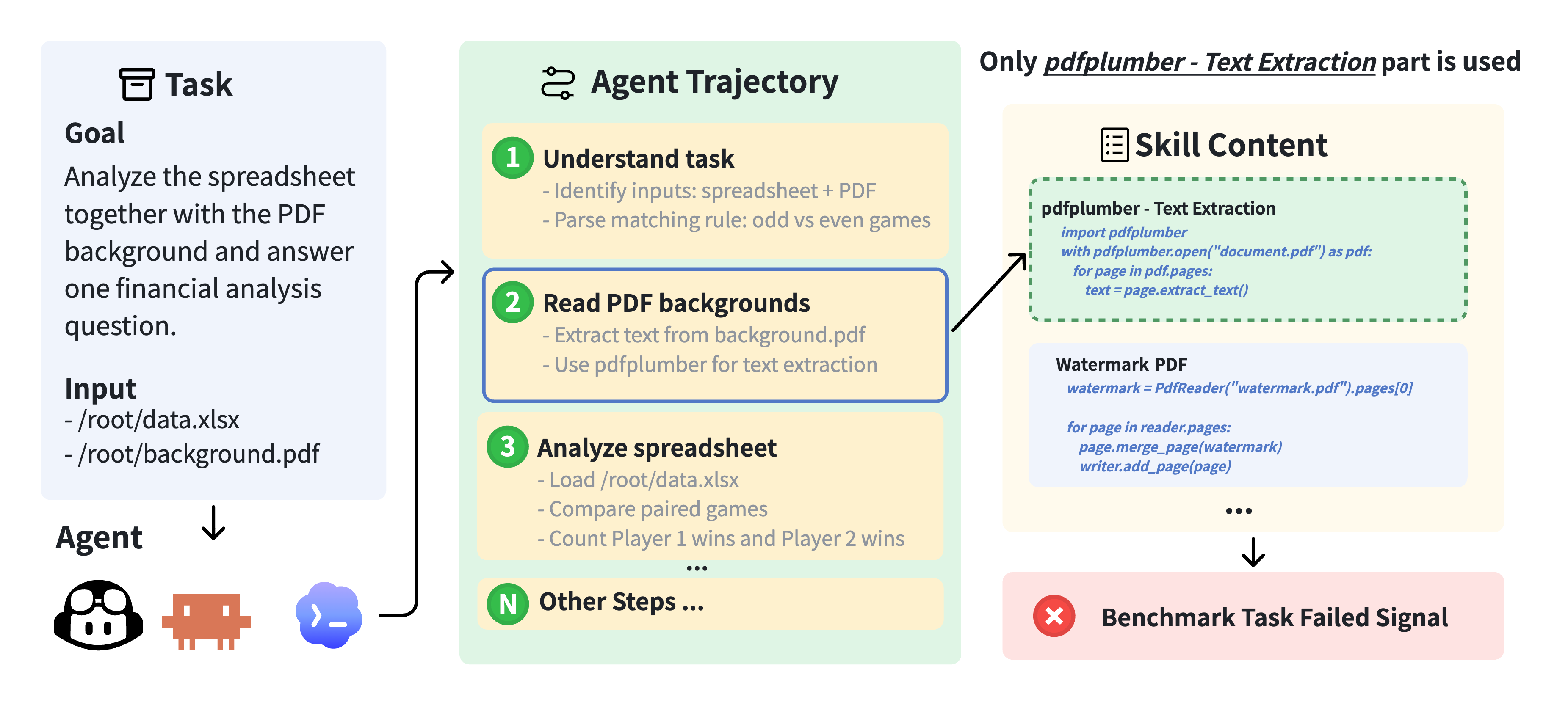}
    \caption{A benchmark task can cover only a fragment of a reusable skill. A task-level outcome is ambiguous even when the task fails: it does not reveal which parts of a reusable skill were exercised, nor whether the agent followed the relevant skill instructions when those parts were exercised.}
    \label{fig:intro-skill-coverage}
\end{figure}

Although current benchmarks demonstrate that skills can improve task performance~\citep{Li2026SkillsBenchBH,Zhong2026SkillLearnBenchBC}, their task-level metrics do not reveal which parts of a reusable skill were exercised, nor whether the agent followed the relevant skill instructions when those parts were exercised.
Figure~\ref{fig:intro-skill-coverage} illustrates this gap: the agent may complete the task by using only the PDF text-extraction part of a skill, leaving other instructions unexercised.

This distinction matters in two ways. First, from a software testing
perspective, a reusable artifact should be adequately tested~\citep{Gao2004TestingComponentBasedSoftware}.
Since skills are intended to be reused across tasks, models, and agents, they
resemble software components whose documented behaviors should be adequately
exercised, not merely associated with a successful task. Second, after a skill
behavior is exercised, task failures still do not reveal whether the skill is ineffective nor agents do not follow the instructions. A task verifier collapses skill-use details into the same
outcome, providing no fine-grained signal about how the agent used the skill.

To provide a more informative signal for skill evaluation, we introduce
\textit{skill coverage}, a trajectory-based test-adequacy metric for reusable
agent skills. The metric is built on \textit{skill behavior constraints}
(\sbcs), which translate natural-language skill instructions into
semi-structured constraints in the Easy Approach to Requirements Syntax
(EARS) style~\citep{mavin2009ears}. Each \sbc specifies an applicability
condition and an expected agent behavior extracted from the skill document.
\begin{center}
\begingroup
\setlength{\fboxsep}{4pt}
\setlength{\fboxrule}{0.4pt}
\fbox{%
\begin{minipage}{0.92\columnwidth}
\small
\textbf{EARS-style \sbc.}
\emph{When adding calculated values to a spreadsheet, the agent shall use Excel formulas to calculate them.}
\end{minipage}}
\endgroup
\end{center}
Given an \sbc, following oracle-based coverage~\citep{schuler2011assessing}, we define when it is \textit{covered} by an execution: the observable trajectory must instantiate the constraint's applicability condition and provide judgeable evidence for the agent behavior to follow the instruction or violate it.
We separate this coverage decision from the test verdict. A constraint is either
\textit{not covered} or \textit{covered}; if it is covered, the evidence receives
a \textit{Pass} verdict when the agent satisfies the expected behavior and a
\textit{Fail} verdict when the agent violates it.
For example, if the agent never adds calculated values to a spreadsheet, the above \sbc is \textit{not covered}. If the trajectory shows that the agent used Excel formulas to calculate values in the spreadsheet, the verdict is \textit{Pass}; if it shows that the agent calculated the values in Python or another tool, the verdict is \textit{Fail}.

Our application to SkillsBench shows why this distinction matters. We first validate
the llm coverage judge against human judgments on sampled
constraint and trajectory pairs, obtaining 88.58\% agreement on the coverage
denominator, where both \textit{Pass} and \textit{Fail} count as covered.
We then apply the validated protocol to SkillsBench. Across five agent--model
configurations from the leaderboard, benchmark executions cover only
38.66--45.51\% of the extracted skill behavior constraints on average, leaving
more than half of the documented skill behaviors without judgeable execution
evidence.
This result shows that task-level benchmark outcomes leave artifact-level
adequacy under-tested: even when a skill is available to the agent, large
portions of its documented behavior may never become observable in benchmark
trajectories.
Then, we apply the coverage judge to failed tasks to analyze why they failed:
whether the skill itself was ineffective or the agent failed to follow the
skill instructions.
We locate \sbcs assigned
a \textit{Fail} verdict: documented skill behaviors that were exercised in the
trajectory but violated by the agent. We use these failed constraints
as diagnostic intervention targets, strengthening the corresponding skill
content only by emphasizing the original instructions that the agent failed to
follow. We then run the same tasks three times under the same verifier with the
strengthened skills. The strengthened skills yield an average recovery rate of
16.0\% across the five agent--model rows.
These results show that skill coverage is both a test-adequacy metric and a
fine-grained signal for observing skill-use behavior. In particular, when a task
fails, failed constraint labels provide actionable evidence for improving agent
skill effectiveness.

This paper makes the following contributions:
\begin{itemize}
    \item \textbf{A skill-level test-adequacy formulation for reusable agent
    skills.}
    We formulate skill evaluation as coverage over
    \textit{skill behavior constraints} (\sbcs), which translate natural-language
    skill instructions into semi-structured constraints with an applicability
    condition and an expected behavior.
    The formulation defines when an agent trajectory covers an \sbc and
    separates the coverage decision from the test verdict: covered behavior
    either satisfies (\textit{Pass}) or violates (\textit{Fail}) the documented
    skill expectation.

    \item \textbf{A validated coverage measurement protocol for SkillsBench.}
    We validate a DeepSeek V4 Flash coverage judge against human labels on
    sampled constraint--trajectory pairs, obtaining 88.58\% agreement on the
    coverage denominator and 81.89\% three-label agreement. Applying the
    validated protocol to five SkillsBench agent--model configurations, we find
    that existing benchmark executions cover only 38.66--45.51\% of the
    extracted \sbcs on average.

    \item \textbf{Evidence that failed constraint labels can diagnose skill-use
    failures and guide strengthening.}
    We use \sbcs assigned a \textit{Fail} verdict in failed tasks as
    diagnostic intervention targets, strengthen the corresponding skill content
    only by emphasizing the original instructions that the agent failed to
    follow, and run the same tasks under the original benchmark verifier. The
    strengthened skills yield an average recovery rate of 16.0\% across the
    five agent--model rows over three strengthened-skill attempts, showing that
    failed constraint labels make skill-use behavior observable and provide
    actionable evidence for improving agent skill effectiveness.
\end{itemize}

\section{Problem Formulation}

Task outcomes alone do not specify which parts of a reusable skill were
exercised, nor whether observed failures provide actionable evidence for
improving agent skill effectiveness. To make this evidence observable, we need
two objects: a finite denominator of documented skill behaviors, and a coverage
relation that says when an agent trajectory covers each behavior constraint.
This section defines both.

\subsection{From Skills to Skill Behavior Constraints}
\label{sec:skill-to-sbcs}

A reusable skill is not merely a one-off prompt fragment for completing a
single task. Rather, it is a natural-language artifact that guides how an
agent should behave across a class of future tasks. Such guidance may describe
what the agent should do, what it should avoid, which workflow steps it should
follow, how it should use a documented tool path, or what properties its
delivered artifacts should satisfy~\citep{Li2026SkillsBenchBH,
Zhong2026SkillLearnBenchBC}. These prescriptions create
\emph{skill behavior constraints} on the agent.

This view is consistent with the role of requirements in requirements
engineering, where requirements capture desired system behaviors and
constraints, and are expected to be verifiable and traceable to their sources
\citep{zave1997four,gotel1994analysis,iso29148}. However, a skill document is
typically not written as a conventional requirements specification. It may mix
instructions, examples, tool usage guidance, workflow steps, common mistakes, and
artifact expectations in free-form natural language. Treating
the entire skill document as a set of behavior constraints would therefore be
too coarse and ambiguous.

We instead define a conservative abstraction called a \emph{Skill Behavior
Constraint} (\sbc). An \sbc translates a natural-language skill instruction into
a semi-structured, EARS-normalized constraint with an applicability condition
and an expected agent behavior. It specifies what the agent shall do, avoid, or
deliver under specific conditions. The purpose of this abstraction is not to
recover the full semantics of the skill document. Rather, it projects the
testable portion of the skill into a finite and auditable set of expected
agent behaviors.

Table~\ref{tab:skill-source-block-types} summarizes how we treat different
skill-source blocks when constructing the audited \sbc set.
It also shows that constructing \sbcs is
selective. We do not convert every part of a skill document into a constraint. A
source block is retained only when it imposes an observable
expectation on the agent, such as an obligation, prohibition, tool-use pattern,
workflow step, or deliverable property. Metadata, background rationale, and
purely descriptive examples are excluded because they do not define behavior
that a trajectory can exercise. Retained blocks are split or tightened when
needed so that each \sbc has one applicability condition and one expected
behavior, while remaining linked to its original source span for auditability.
We denote the resulting audited set for skill $S$ as $\mathcal{C}(S)$.

\begin{center}
\begingroup
\setlength{\fboxsep}{4pt}
\setlength{\fboxrule}{0.4pt}
\fbox{%
\begin{minipage}{0.92\columnwidth}
\small
\textbf{Normalized \sbc form.}
\[
\begin{aligned}
&\text{When } \langle \text{applicability condition} \rangle,\\
&\text{the agent shall }
\langle \text{observable agent behavior} \rangle.
\end{aligned}
\]
\end{minipage}}
\endgroup
\end{center}

The \emph{When} clause states when the constraint is applicable, and the
\emph{shall} clause states the expected agent behavior. This normalization is
deliberately lightweight: it defines the testable portion of a skill document
without treating the whole skill as a complete requirements specification.

\begin{table*}[t]
\centering
\caption{Skill-source block treatment for constructing the audited \textsc{SBC} set.}
\label{tab:skill-source-block-types}
\footnotesize
\setlength{\tabcolsep}{4pt}
\renewcommand{\arraystretch}{1.08}
\newcommand{\producesconstraint}{\(\checkmark\)\,\textbf{Yes.}}
\newcommand{\doesnotproduceconstraint}{\(\times\)\,\textbf{No.}}
\begin{tabularx}{\textwidth}{@{}
  >{\raggedright\arraybackslash}p{0.24\textwidth}
  >{\raggedright\arraybackslash}p{0.28\textwidth}
  >{\raggedright\arraybackslash}X@{}}
\toprule
\textbf{Source block type} &
\textbf{Example source block} &
\textbf{Constraint decision and normalized form} \\
\midrule
\multicolumn{3}{@{}l}{\textit{Excluded from the audited SBC set}} \\
\addlinespace[1pt]
\texttt{document\_metadata} &
\texttt{Author: A. Smith}; \texttt{Version: 1.2}; \texttt{Updated: 2026-03-01} &
\doesnotproduceconstraint{} It records provenance or maintenance information, not behavior to be tested. \\
\addlinespace[2pt]
\texttt{background} &
\emph{``This skill was created to keep related workflow notes in one place.''} &
\doesnotproduceconstraint{} Document-level motivation only; it does not prescribe, prohibit, or constrain observable agent behavior. \\
\midrule
\multicolumn{3}{@{}l}{\textit{Retained as candidate sources for audited SBCs}} \\
\addlinespace[2pt]
\texttt{agent\_behavior\_constraint} &
\emph{``Always use Excel formulas instead of calculating values in Python and hardcoding them.''} &
\producesconstraint{}
When adding calculated values to a spreadsheet, the agent shall use Excel formulas instead of calculating values in Python and hardcoding them. \\
\addlinespace[4pt]
\texttt{deliverable\_constraint} &
\emph{``Every Excel model MUST be delivered with ZERO formula errors.''} &
\producesconstraint{}
When delivering an Excel model, the agent shall ensure the workbook contains zero formula errors. \\
\addlinespace[4pt]
\texttt{operation\_unit} &
\emph{``Analyze which funds hold a stock to the most extent''}; \texttt{python scripts/analysis.py ...} &
\producesconstraint{}
When analyzing which funds hold a particular stock to the most extent, the agent shall use the analysis.py script. \\
\addlinespace[4pt]
\texttt{workflow\_constraint} &
\emph{``Build a layer inventory before extracting geometry. Include entity counts and entity types per layer.''} &
\producesconstraint{}
When the DXF has been opened and inspected, the agent shall build a layer inventory before extracting geometry, including entity counts and entity types per layer. \\
\addlinespace[4pt]
\texttt{common\_mistake} &
\emph{``Never impose standardized formatting on files with established patterns.''} &
\producesconstraint{}
When modifying a spreadsheet with established formatting patterns, the agent shall not impose standardized formatting that conflicts with the existing file conventions. \\
\bottomrule
\end{tabularx}
\end{table*}

\subsection{Skill Behavior Constraint Coverage}
\label{sec:sbc-coverage}

Given the audited set $\mathcal{C}(S)$, coverage asks which documented skill
constraints are covered by an agent trajectory. A task-level oracle
verdict does not by itself characterize skill test adequacy: \textbf{loading a skill
into an agent context, or having the agent recall skill content internally, does
not mean that any particular documented behavior has been exercised.} We
therefore define coverage by the applicability condition of an \sbc. Once the
condition is instantiated, the corresponding \emph{shall} clause is under test.
The resulting verdict is \emph{Pass} if the agent follows the expected behavior
and \emph{Fail} if the agent violates or omits it.

For example, consider the \sbc:
\emph{When adding calculated values to a spreadsheet, the agent shall use Excel
formulas to calculate them.} A trajectory covers this constraint when the agent
enters the applicability condition by adding calculated values to a spreadsheet.
If the agent merely loads the skill
or discusses the instruction without actually action that adding calculated values to a
spreadsheet, the constraint is not covered and no verdict is issued.
The verdict is \emph{Pass} only if the observable agent actions show that the
agent used Excel formulas for those values, such as by writing formulas into the
workbook. If the trajectory instead shows that the agent computed the values in
Python, wrote static values, or otherwise produced no action showing the use of
Excel formulas, the verdict is \emph{Fail}. 

Let $S$ be a skill document and let $\mathcal{C}(S)$ be the audited set of
\sbcs extracted from it. A constraint $c \in \mathcal{C}(S)$ contains an
applicability condition $C_c$ and an expected behavior $B_c$. Let $\tau$ be one
agent trajectory, and let $\mathsf{Obs}(\tau)$ denote
the observable record of that trajectory, such as llm responses, tool calls, and artifact outputs.
For
a constraint--trajectory pair $(c,\tau)$, we define coverage directly:
\[
\mathsf{covered}(c,\tau) = 1
\quad\text{if}\quad
\mathsf{Obs}(\tau) \text{ instantiates } C_c.
\]
When $\mathsf{covered}(c,\tau)=1$, the pair receives one of two verdicts:
\[
\mathsf{verdict}(c,\tau)=
\begin{cases}
\mathsf{Pass}, & \text{if } \mathsf{Obs}(\tau) \text{ satisfies } B_c,\\
\mathsf{Fail}, & \text{if } \mathsf{Obs}(\tau) \text{ violates or omits } B_c.
\end{cases}
\]

\section{Method}
\label{sec:method}
\subsection{Overview}
Our method maps skill documents and observable agent trajectories to
constraint-level labels that support both adequacy measurement and skill
strengthening. Given a benchmark task, its task-associated skill documents, and
the trajectory produced by an agent, the procedure produces two artifacts: an
audited set of source-grounded skill behavior constraints and a set of
trajectory-grounded labels for those constraints.

The method has three operational stages. First, \emph{constraint extraction} turns a
natural-language skill document into an auditable denominator of \sbcs.
Second, \emph{trajectory labeling} compares each \sbc against observable
execution evidence and assigns one of three outcomes: \emph{not covered},
\emph{Pass}, or \emph{Fail}. Third, \emph{failure-guided strengthening} treats
\emph{Fail} outcomes as
candidate skill-following signals, selects plausible original
skill content to emphasize, and runs the same tasks with strengthened skills.
This decomposition separates the three questions that the paper evaluates:
whether the LLM labels are reliable enough to use at scale, how much of the
skill denominator current benchmark tasks cover, and whether failed labels are
useful for diagnosing whether agents follow relevant skill instructions.

\subsection{Extracting Skill Behavior Constraints}
\label{sec:method-extract}

The extraction stage implements the \sbc boundary from
Section~\ref{sec:skill-to-sbcs} through LLM-assisted candidate extraction
followed by human audit and repair. We use the LLM as a recall-oriented parser
over skill prose, and use human audit to decide which candidates enter the
final \sbc set.

Before running extraction at scale, human annotators first randomly sample a
small set of skill cases and manually write reference extractions for them. For
each sampled case, the annotation marks the relevant source span, decides
whether the text should yield an \sbc, and writes the corresponding
applicability condition and expected behavior. These human-labeled examples are
then used as few-shot demonstrations in the extraction prompt. Given a new skill
document $S$, we provide the full skill document and these demonstrations to the
LLM and ask it to extract candidate \sbcs directly. The prompt defines the \sbc
boundary from Section~\ref{sec:skill-to-sbcs}, requires EARS-style
condition-action output, and uses the demonstrations to illustrate the
source-block categories and inclusion/exclusion decisions in
Table~\ref{tab:skill-source-block-types}. Thus,
Table~\ref{tab:skill-source-block-types} serves as prompt guidance for what
kinds of skill text should or should not produce constraints, rather than as a
separate preprocessing classifier over segmented blocks.

The LLM proposes candidate \sbcs from the full skill document. Each
candidate records the supporting source span, applicability condition,
and expected behavior.
After the LLM extraction, human auditors review and repair the candidate
set before it becomes $\mathcal{C}(S)$. The audit checks whether each candidate
is source-grounded, constraint-bearing, and observable. Auditors
discard candidates derived from background or incidental implementation detail,
merge duplicate candidates, split compound candidates, repair applicability
conditions, and make the expected behavior explicit. For example, a workflow
sentence that says to inspect a file before modifying it can yield a constraint
about the inspection step, while a rationale sentence explaining why the
workflow is useful is removed unless it independently imposes an observable
obligation.

The audited candidates are finally normalized into EARS-style condition-action
forms. This normalization is a control layer over natural-language skill prose,
not a claim that skills are complete formal specifications. Scope-wide
constraints are written with the skill's task context as their condition, while
local constraints introduce additional triggers such as user requests, artifact
states, workflow states, or the agent's choice of a documented tool path. The
output of this stage is the audited set $\mathcal{C}(S)$ used as the
denominator for skill coverage.

\subsection{Labeling Constraints in Agent Trajectories}
\label{sec:method-judge}

The trajectory-labeling stage assigns a label to each constraint--trajectory
pair. The input is a constraint with applicability condition $C_c$, expected
behavior $B_c$, and an observable trajectory $\tau$. We first construct
$\mathsf{Obs}(\tau)$ by
retaining evidence available to an external evaluator: the user request,
visible agent messages, tool calls, tool outputs, command invocations,
generated code, file diffs, artifact states, logs, and final responses.

Because trajectory labeling requires semantic attribution over execution traces,
we use LLM-based coverage attribution only after validating it against human
judgments. Following validation-before-scale practice in prior LLM-as-a-judge
evaluations, where automatic judges are compared with human or expert judgments
before broader use~\citep{Zheng2023JudgingLW,Liu2023GEvalNE,dubois2024lengthcontrolled},
we first draw a validation subset from the constraint--trajectory population
using a predefined sample-size plan rather than a convenience
set~\citep{krejcie1970determining}. The sample preserves variation across skills,
tasks, and preliminary coverage outcomes when available.

We apply the same labeling rubric in two agreement checks. First, two human
annotators independently label the sampled constraint--trajectory pairs to test
whether the rubric is reproducible for human evaluators. Second, after this
human--human check, we run the LLM judge on the same sampled pairs and compare
its labels with the human annotations in both the original three-label space and
the coverage denominator that counts both \emph{Pass} and \emph{Fail} as
covered. This estimates whether the judge reproduces the intended coverage
criterion. Only after this human--LLM
agreement check do we freeze the judging prompt, evidence format, and label
mapping, and then apply the same fixed protocol to the benchmark-scale trace
set. This design follows fine-grained rubric and criterion-based evaluations,
where LLM judges apply predefined criteria at scale after calibration or
validation~\citep{Ye2023FLASKFL,ICLR2024_80348535,Starace2025PaperBenchEA,Arora2025HealthBenchEL}.
Thus, the coverage labels are outputs of a validated measurement protocol, not
uncalibrated model opinions.

All human and LLM labels use the same two-step decision procedure. First, the
rubric checks whether the observable task context and trajectory instantiate
$C_c$. If the condition is not instantiated, the pair is \emph{not covered}. If
the condition is instantiated, the pair is covered and receives a verdict:
\emph{Pass} if the observed behavior satisfies $B_c$ and \emph{Fail} if it
violates or omits $B_c$. A final response that merely claims compliance is not
sufficient unless the final response itself is the object constrained by the
\sbc. The judgment must be grounded in concrete trace evidence such as a tool
call, file state, command output, generated artifact, or cited response content.

\begin{figure*}[t]
\centering
\includegraphics[width=\textwidth]{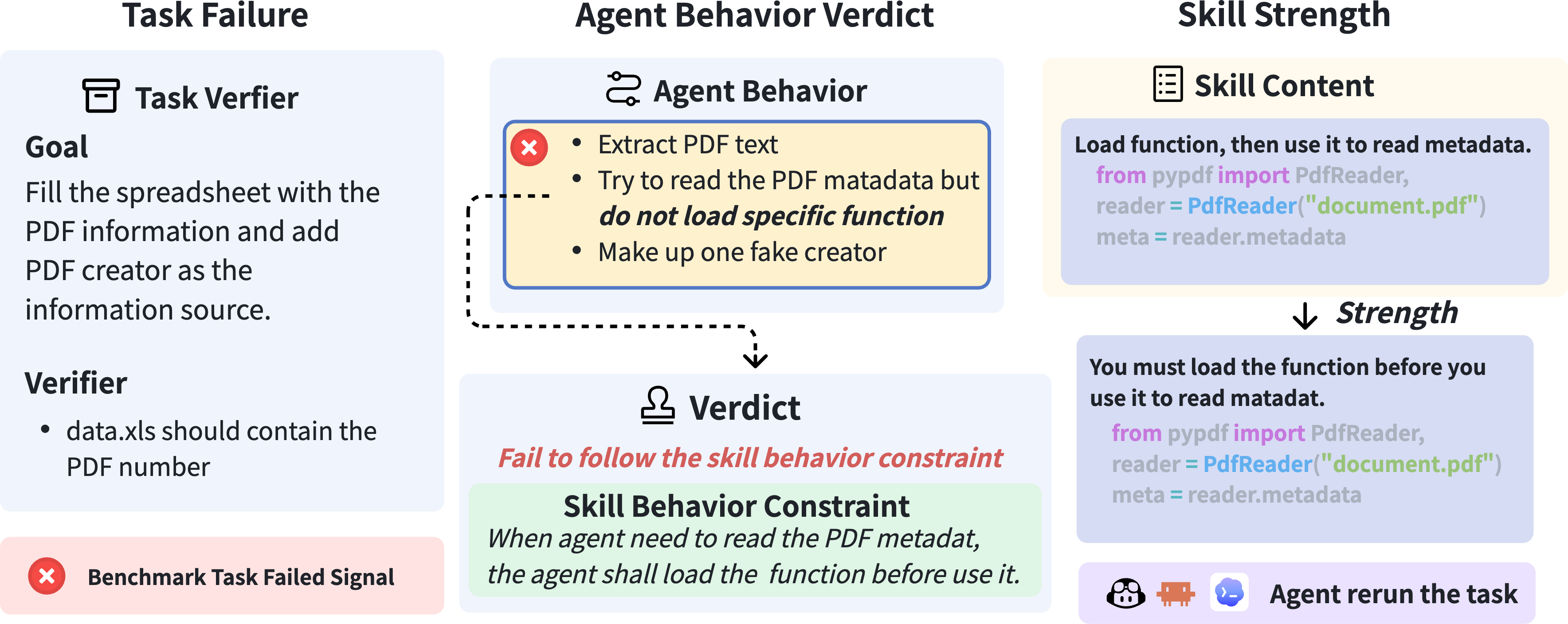}
\caption{Example failure-guided skill strengthening. A failed PDF task first
exposes only a benchmark failure. The coverage judge links the trajectory to an
applicable \sbc and labels the behavior as \emph{Fail}: the agent tried to read
PDF metadata without loading the reader path specified by the skill.
Strengthening only emphasizes that original reader-loading instruction before
the same task is run again.}
\label{fig:strength-skill}
\end{figure*}
\subsection{Using Failed Constraints for Skill Strengthening}
\label{sec:method-strengthening}

A \emph{Fail} verdict is useful because it says more than the task verifier
alone. A failed constraint was exercised, but the observable trajectory shows
that the agent violated or omitted the required skill behavior. In failed tasks,
these labels therefore expose a skill-use situation that the task verifier alone
cannot observe: the skill may not provide effective guidance for the task, or
the agent may have failed to follow relevant guidance that was already present
in the skill.

For each failed benchmark task, we collect the constraints that receive a
\emph{Fail} verdict and ask a separate triage judge to review the task
instruction, failed \sbc, skill-use context, and final agent response to decide
whether the violated skill behavior plausibly contributed to the task failure.
The judge labels each candidate as $\mathsf{Strong}$,
$\mathsf{Partial}$, or $\mathsf{Weak}$. This triage step decides whether
the failed skill behavior is worth using as an instruction-emphasis signal.
The strengthening step itself is deliberately simple. For strong and partial
candidates, the prompt receives the original skill and the trajectory-derived
failure patterns, then rewrites the relevant skill content only to foreground
the original instruction that the agent failed to follow. The edit does not add
a new task solution or a new verifier target; it makes the existing instruction
more salient by turning it into an explicit checklist, ordering constraint,
do-not-skip verification gate, or concrete-evidence requirement.

Figure~\ref{fig:strength-skill} illustrates this workflow with a PDF task. The
original PDF skill already describes the expected metadata-reading path:
construct a \texttt{PdfReader} for the document and then read
\texttt{reader.metadata}. The failed trajectory attempts to read the metadata
but does not load the required reader path, so the corresponding \sbc receives a
\emph{Fail} verdict. The strengthened skill then foregrounds the original
instruction by making the reader-loading step explicit before the same task is
run with the strengthened skill. This edit does not provide the spreadsheet
answer, introduce a new
metadata extraction strategy, or use verifier pass as a training target.
We then run the same task with the strengthened skill and score the result
only with the original benchmark verifier. A recovered task outcome suggests
that emphasizing the original instruction changed agent behavior; a remaining
failure suggests that the agent still did not follow the emphasized instruction
or that other task-level causes dominated. The protocol therefore uses failed
constraints as a controlled instruction-following probe, not as supervised
training from successful solutions.

\section{Evaluation}
\label{sec:evaluation}

We evaluate the framework through three research questions. RQ1 asks whether
the LLM judge can reproduce human trajectory labels. RQ2 applies the validated
judge at SkillsBench scale and estimates how much of the audited \sbc
denominator is covered by current benchmark tasks. RQ3 tests whether
\emph{Fail} verdicts are useful beyond measurement by using them to
guide skill strengthening and run previously failed tasks with strengthened
skills. The evaluation therefore separates label reliability, benchmark-scale
coverage, and label-guided intervention.

\subsection{Experimental Setup}

\subsubsection{Benchmark}
We apply \emph{skill coverage} to SkillsBench~\citep{Li2026SkillsBenchBH}, a
recent benchmark for measuring whether agent skills improve task completion.
SkillsBench contains 87 source tasks across diverse domains, which are
associated with 202 task-linked skill artifacts in our coverage slice. The
benchmark pairs tasks with curated skills and deterministic verifiers, and its
primary comparison asks whether access to those skills improves task pass rate.
The paired skills are not written as per-task solution notes: SkillsBench builds
its skill pool from public repositories, community marketplaces, and partner
contributions, deduplicates skills by \texttt{SKILL.md} content hash, and
selects benchmark skills from a quality-scored top quartile using criteria such
as completeness, clarity, specificity, and examples. Thus, the benchmark gives
us task-associated skills that are intended to be high-quality reusable
procedural artifacts rather than task-specific hints.
This outcome-level design makes SkillsBench a useful setting for our
complementary question: once a skill is available during a benchmark execution,
which documented behaviors in the skill become observable in the agent
trajectory, and when they are observable, does the agent satisfy or violate
them? Because the paired skills are intended as reusable procedural artifacts,
task success and skill adequacy are meaningfully separable: an agent may
complete a task while exercising only a subset of the behaviors documented by
the associated skills.
We therefore report benchmark-scale coverage on 87 source tasks with
202 task-associated skills for which we collected task outcomes, observable
agent trajectories, and skill-level coverage judgments.

\subsubsection{Agent and Model Selection}
We evaluate five agent--model configurations. They cover three
agent harnesses (Codex, Claude Code, and
OpenHands)~\citep{openai2026codex,anthropic2026claudecode,Wang2024OpenHandsAO}
and four model families (GPT-5.5, Claude Opus 4.7, DeepSeek V4, and
GLM 5.1)~\citep{openai2026gpt55systemcard,
anthropic2026claudeopus47systemcard,deepseekai2026deepseekv4,
zai2026glm51,glm5team2026glm5}. To keep coverage-judging cost bounded while
avoiding the selection bias of uniformly sampling a few rows at random, we
stratified rows by their average task pass rate across three reruns of the same
87 source tasks. This stratification improves result reliability by ensuring
that the retained slice covers different baseline success regimes rather than
only the configurations that happen to be sampled. We use three tiers: high
($\geq 50\%$), middle (35--50\%), and low ($<35\%$). We selected
representatives from all three tiers: two
high-pass-rate configurations
(Codex with GPT-5.5, 53.26\%; and Claude Code with Opus 4.7, 52.87\%), two
middle configurations (OpenHands with DeepSeek V4 Pro, 40.23\%; and OpenHands
with GLM 5.1, 37.55\%), and one low-pass-rate configuration (OpenHands with
DeepSeek V4 Flash, 32.95\%). This design keeps the benchmark-scale coverage
analysis to five configurations while still covering different agent/model
families. Each configuration is evaluated on the same 87 source tasks with
three reruns per task. We use the benchmark verifier to obtain task-level
success outcomes.

\subsubsection{Coverage Judge Model}
To control the cost of coverage analysis, we use DeepSeek V4 Flash as the LLM
constraint judge. It receives only the \sbc{} and the whole observable
trajectory, and applies the two-step trajectory-labeling rubric from
Section~\ref{sec:method-judge}. The judge assigns one of three labels:
\emph{not covered}, \emph{Pass}, or \emph{Fail}.

\subsection{Metrics}

\subsubsection{Task Pass Rate}
We use the benchmark verifier to measure task-level success. For each
agent--model configuration, the task pass rate is the average pass rate across
three reruns of the same 87 source tasks. We use this metric to stratify
configurations and to condition coverage by task outcome. It is intentionally
kept separate from skill coverage: a task can pass without exercising every
documented behavior in the associated skills.

\subsubsection{SBC Labels and Coverage}
The coverage judge assigns each constraint--trajectory pair one of three
labels: \emph{not covered}, \emph{Pass}, or \emph{Fail}. A constraint is
\emph{not covered} when the trajectory does not instantiate its applicability
condition. A \emph{Pass} label means that the trajectory covers the constraint
and satisfies the required behavior, while a \emph{Fail} label means that the
trajectory covers the constraint but violates or omits the required behavior.
For benchmark-scale coverage reporting, both \emph{Pass} and \emph{Fail}
contribute to the covered numerator:
\[
\mathrm{Coverage}(\mathcal{J}) =
\frac{|\{(c,\tau)\in\mathcal{J}: \ell(c,\tau)\in
\{\mathsf{Pass},\mathsf{Fail}\}\}|}{|\mathcal{J}|},
\]
where $\mathcal{J}$ is a set of constraint--trajectory judgments and
$\ell(c,\tau)$ is the assigned label. Under this definition, \emph{Fail}
counts as covered because the relevant skill behavior was exercised, even
though the agent did not satisfy it. Conversely, an uncovered constraint is not
a violation; it means that the current execution did not test that behavior.

\subsubsection{Aggregation}
For configuration-level coverage, we compute the coverage rate for each rerun and
report the mean and range across the three reruns. For per-skill coverage, we
merge the three reruns for each task-associated skill artifact and compute the
covered fraction over the merged constraint--trajectory judgments. For
task-outcome-conditioned coverage, we group task executions by the verifier
outcome and compute coverage over the constraints associated with executions in
each group.

\subsection{RQ1: LLM Judge Reliability}
\label{sec:measurement-validation}
RQ1 asks whether the trajectory-labeling rubric is reliable enough to support
benchmark-scale coverage measurement. Across the current 87-task SkillsBench
scope, our extraction and audit produce 4,283 unique \sbcs{} from the 202
task-associated skill artifacts. Because some skills are reused by multiple
tasks, expanding these constraints to task-associated skill instances yields
5,298 task-associated \sbc{} instances for one rerun. For judge validation, we
pair each instance with the corresponding Codex + GPT-5.5 trajectory, which comes from the highest-pass-rate configuration. Following a
statistically principled sampling method at a 95\% confidence
level~\citep{krejcie1970determining}, this population gives a required sample
size of 359 cases.

Two human auditors independently label each sampled constraint--trajectory pair
as \emph{not covered}, \emph{Pass}, or \emph{Fail}. We first measure human--human agreement to check that the
rubric is reproducible. When \emph{Pass} and \emph{Fail} are both counted as
covered, the two human annotators agree on 352/359 cases, for an exact
agreement of 98.05\% and Cohen's $\kappa=0.961$. In the original three-label
space, they agree on
343/359 cases, for an exact agreement of 95.54\% and $\kappa=0.924$. These
results indicate that the label space is stable enough for judge validation
rather than being inherently ambiguous.
We then validate the initial DeepSeek V4 Flash judge against the human labels
on the same sample. For the same coverage denominator, the judge agrees with
the human audit on 318 cases, for an exact agreement of 88.58\% and Cohen's
$\kappa=0.770$. In the original three-label space, it agrees on 294 cases, for
an exact agreement of 81.89\% and $\kappa=0.695$.

\begin{tcolorbox}[colback=gray!5, colframe=black!60, boxrule=0.5pt, arc=2pt, left=6pt, right=6pt, top=4pt, bottom=4pt]
\textbf{Answer to RQ1.} The DeepSeek V4 Flash judge reproduces human coverage labels with high reliability. The rubric is stable enough to support benchmark-scale coverage measurement, and the judge can be used to identify failed constraints for coverage-guided skill strengthening.
\end{tcolorbox}

\begin{figure*}[t]
\centering
\includegraphics[width=\textwidth]{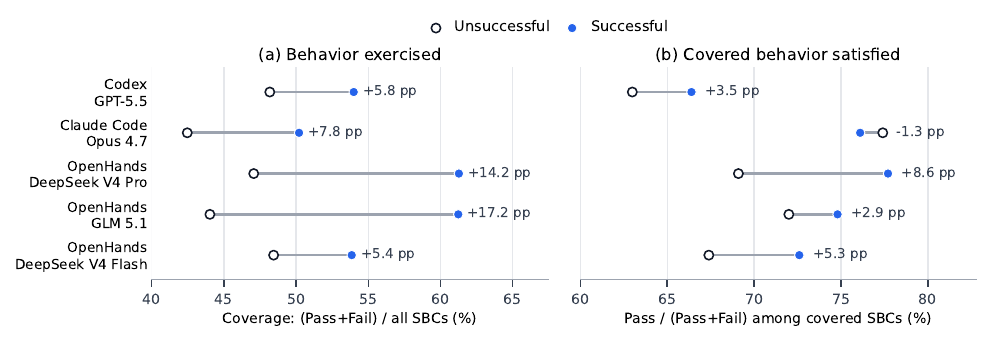}
\caption{Task-outcome-conditioned skill behavior exercise and satisfaction.
Panel (a) reports coverage, counting both \emph{Pass} and \emph{Fail} as
covered. Panel (b) reports \emph{Pass}/(\emph{Pass}+\emph{Fail}) among covered
constraints. Open points are unsuccessful executions, filled points are
successful executions, and labels show the success-minus-failure gap.}
\label{fig:task-outcome-coverage}
\end{figure*}

\subsection{RQ2: SkillsBench Coverage}

For RQ2, we estimate benchmark-scale coverage from repeated executions rather
than from a single trajectory. For each retained agent--model configuration, we
rerun the same 87 source tasks three times, yielding three trajectory slots per
configuration. We then judge each slot independently: every task-associated
\sbc{} instance is paired with the corresponding trajectory, labeled by the
DeepSeek V4 Flash judge, and counted as covered when the label is either
\emph{Pass} or \emph{Fail}. We compute coverage separately for each slot and
report the mean and min--max range across the three slots. This protocol reduces
sensitivity to idiosyncratic single-run behavior and makes the reported
coverage values reflect a small repeated-trajectory sample.

\begin{table}[!ht]
\centering
\caption{Mean task success and SBC coverage across three reruns.}
\label{tab:skillsbench-coverage-summary}
\footnotesize
\renewcommand{\arraystretch}{1.08}
\begin{tabular}{@{}llcc@{}}
\toprule
\textbf{Agent} &
\textbf{Model} &
\textbf{\shortstack{Task success}} &
\textbf{\shortstack{SBC coverage}} \\
\midrule
Codex & GPT-5.5 & 53.26\% & 43.60\% \\
Claude Code & Opus 4.7 & 52.87\% & 38.66\% \\
OpenHands & DeepSeek V4 Pro & 40.23\% & 45.51\% \\
OpenHands & GLM 5.1 & 37.55\% & 41.64\% \\
OpenHands & DeepSeek V4 Flash & 32.95\% & 43.73\% \\
\bottomrule
\end{tabular}
\end{table}

Table~\ref{tab:skillsbench-coverage-summary} reports the basic row-level
results: task success and \sbc coverage for each agent--model configuration,
averaged across the three reruns. The first pattern is that coverage is low for
every retained row: average coverage remains between 38.66\% and 45.51\%,
leaving 54.49--61.34\% of the audited skill-behavior denominator untested by
these executions. The second pattern is that row-level task pass rate is not
aligned with row-level coverage. Codex with GPT-5.5 and Claude Code with Opus
4.7 have similar task success rates (53.26\% and 52.87\%) but different average
coverage (43.60\% and 38.66\%). Conversely, OpenHands with DeepSeek V4 Pro has
the highest average coverage (45.51\%) despite a middle task success rate
(40.23\%), while OpenHands with DeepSeek V4 Flash has the lowest task success
rate (32.95\%) but still reaches 43.73\% average coverage. Thus, configuration-
level task success does not explain how much of the skill artifact is exercised;
\sbc coverage is measuring a different adequacy property.

Figure~\ref{fig:task-outcome-coverage} adds two task-outcome-conditioned views
of the same judgments and shows where task outcome does matter. Panel (a)
shows that successful executions usually traverse more skill-relevant behavior:
they cover more of the associated \sbc denominator than unsuccessful executions
in all five rows, with pass-minus-fail gaps from 5.42 to 17.22 percentage
points and an average gap of 10.09 percentage points. However, successful
executions still cover only 50.21--61.29\% of the denominator on average, while
unsuccessful executions cover 42.46--48.44\%. Passing a task therefore tends to
exercise more documented skill behavior, but it still leaves large parts of the
skill artifact unobserved. Panel (b) examines only the covered portion of the
denominator: among constraints that are covered, successful task executions have
a 73.3\% \emph{Pass} rate, compared with 69.5\% for unsuccessful executions.
This satisfaction gap is modest (+3.8 percentage points overall) and varies by
row, from -1.3 percentage points for Claude Code with Opus 4.7 to +8.6
percentage points for OpenHands with DeepSeek V4 Pro.

\begin{tcolorbox}[colback=gray!5, colframe=black!60, boxrule=0.5pt, arc=2pt, left=6pt, right=6pt, top=4pt, bottom=4pt]
\textbf{Answer to RQ2.}
Existing SkillsBench executions exercise nontrivial parts of the skill
denominator, but still cover less than half of the extracted skill behaviors.
Successful tasks typically exercise more skill-relevant behavior.
\end{tcolorbox}

\subsection{RQ3: Coverage-Guided Skill Strengthening}
\label{sec:rq3-skill-strengthening}

We first inspect the coverage
analysis for each failed task and retain
constraints that receive a \emph{Fail} verdict.
The judge assigns one of three labels: $\mathsf{Strong}$ if the failed
constraint is a clear original-instruction emphasis target for the task
failure, $\mathsf{Partial}$ if the failed constraint is plausibly relevant but
other failure causes may also matter, and $\mathsf{Weak}$ if the constraint
failure does not provide a credible instruction-emphasis target. We strengthen
only $\mathsf{Strong}$ and $\mathsf{Partial}$ cases, run the same task with the
strengthened skill, and count task success only from the original verifier
reward. The strengthening prompt receives the full original skill and the
trajectory-derived failure patterns. It only emphasizes the original skill
instructions that the failed trajectory did not follow.
We run each strengthened task three times under the original benchmark verifier
and report the average strengthened success count.

\begin{table}[ht]
    \centering
    \caption{Coverage-guided skill strengthening results.}
    \label{tab:rq3-skill-strengthening}
    \footnotesize
    \setlength{\tabcolsep}{1pt}
    \renewcommand{\arraystretch}{1.08}
    \begin{tabular*}{\columnwidth}{@{\extracolsep{\fill}}lccc@{}}
        \toprule
        \textbf{Agent--model} &
        \textbf{\shortstack{Failed tasks}} &
        \textbf{\shortstack{Success avg.}} &
        \textbf{\shortstack{Recovery rate}} \\
        \midrule
        Codex / GPT-5.5                   & 38 & 11.7 & 30.7\% \\
        Claude Code / Opus 4.7            & 46 & 8.3  & 18.1\% \\
        OpenHands / DeepSeek V4 Pro       & 57 & 9.0  & 15.8\% \\
        OpenHands / GLM 5.1               & 54 & 4.7  & 8.6\%  \\
        OpenHands / DeepSeek V4 Flash     & 57 & 4.0  & 7.0\%  \\
        \bottomrule
    \end{tabular*}
    \vspace{0.25\baselineskip}
\end{table}

Table~\ref{tab:rq3-skill-strengthening} reports an exploratory
coverage-guided strengthening study after one diagnostic baseline execution per
agent--model row. The diagnostic executions contain 46, 38, 57, 57, and 54
failed tasks. The success-average column reports the mean number of successful
task outcomes across the three strengthened-skill attempts.
Recovery rate divides the underlying three-attempt total by the failed
diagnostic tasks times three. Across the five retained rows, the recovery rates
average 16.0\%. The repeated strengthened attempts reduce sensitivity to a
single stochastic trajectory.

Table~\ref{tab:rq3-strengthened-constraint-outcomes} then shifts the analysis
from task outcomes to the strengthened constraints themselves. After rerunning
the tasks with strengthened skills, we re-judge the selected failed constraints
against the new trajectories. The table reports how many constraints were
strengthened and, among selected constraints that are covered again in the
strengthened trajectories, what fraction now receive passing judgments. Covered
pass rate ranges from 43.8\% to 66.2\%.
The Codex and Claude Code rows have the highest covered-pass rates, while the
OpenHands rows cluster around 44\%. This constraint-level view qualifies the
task-level recovery result: strengthening can recover task successes.

Taken together, Tables~\ref{tab:rq3-skill-strengthening}
and~\ref{tab:rq3-strengthened-constraint-outcomes} suggest that in some failed tasks, skill artifacts also
contain useful operational knowledge, but agents do not always instantiate that
knowledge in their trajectories. Because the strengthening prompt only
re-emphasizes original skill instructions, the recovered successes are better
interpreted as improved use of existing skill knowledge rather than new task
knowledge injected by the experiment.
Our \emph{Fail} verdicts of skill behavior constraints provide a fine-grained view of knowledge-use failures in agent trajectories which could be used to enhance the effectiveness of skill artifacts.

\begin{tcolorbox}[colback=gray!5, colframe=black!60, boxrule=0.5pt, arc=2pt, left=6pt, right=6pt, top=4pt, bottom=4pt]
\textbf{Answer to RQ3.}
The strengthening
study shows that skill artifacts can contain useful operational knowledge even
when agents fail the task. The \emph{Fail} verdicts provide a fine-grained signal to distinguish wether the skill is ineffective or the agent is not using it properly. This signal could be used to improve the effectiveness of skill artifacts and guide skill strengthening.
\end{tcolorbox}

\begin{table}[ht]
    \centering
    \caption{Constraint-level outcomes after skill strengthening.}
    \label{tab:rq3-strengthened-constraint-outcomes}
    \footnotesize
    \setlength{\tabcolsep}{1pt}
    \renewcommand{\arraystretch}{1.08}
    \begin{tabular*}{\columnwidth}{@{\extracolsep{\fill}}lcc@{}}
        \toprule
        \textbf{Agent--model} &
        \textbf{\shortstack{Strengthened SBCs}} &
        \textbf{\shortstack{Covered-pass rate}} \\
        \midrule
        Codex / GPT-5.5                   & 94  & 66.2\% \\
        Claude Code / Opus 4.7            & 79  & 63.5\% \\
        OpenHands / DeepSeek V4 Pro       & 171 & 44.9\% \\
        OpenHands / GLM 5.1               & 39  & 44.3\% \\
        OpenHands / DeepSeek V4 Flash     & 62  & 43.8\% \\
        \bottomrule
    \end{tabular*}
\end{table}

\section{Discussion}
\label{sec:discussion}

\subsection{From Coverage Labels to Skill Maintenance}

The practical value of skill coverage is that it turns benchmark traces into a
maintenance worklist for reusable skills. Each label suggests a different next
action. A \emph{not covered} constraint points to a skill behavior that the
current benchmark never tests; the appropriate response is to add or vary tasks
so that the condition, workflow branch, prohibition, or deliverable requirement
actually becomes observable. A \emph{Pass} constraint provides evidence that the
documented behavior can be followed under the observed task conditions. A
\emph{Fail} constraint is a stronger maintenance signal: the condition became
applicable, but the agent violated or omitted the documented behavior in the
trace.

This distinction matters because failed tasks do not all imply the same repair.
Some failures indicate missing skill knowledge: the skill simply does not tell
the agent how to handle the situation. Other failures indicate skill-use
breakdown: the skill already contains useful operational knowledge, but the
agent does not reliably instantiate it. Failed \sbcs make the second case
visible. They identify the exact instruction whose salience, ordering, evidence
requirement, or interaction with the agent harness should be inspected before
rewriting the skill more broadly.

RQ3 illustrates this maintenance use case. The strengthening prompt does not add
successful answers, new task-specific solution knowledge, or verifier feedback
as a training target. It only re-emphasizes original skill instructions that
failed trajectories did not follow. The 16.0\% average recovery rate across
repeated strengthened attempts, together with 43.8--66.2\% covered-pass rates
for selected constraints in the strengthened trajectories, suggests that some
failures occur because agents do not reliably use knowledge already present in
the skill artifact. In such cases, skill coverage helps decide whether the next
maintenance step should be to expand benchmark coverage, clarify an existing
instruction, make an ordering or evidence gate explicit, or investigate why the
agent harness fails to surface the relevant skill content at the right moment.

\subsection{Skill Coverage vs. Code Coverage}

Skill coverage is analogous to code coverage, but the analogy is deliberately
imperfect. If the foundation model is the processor, the agent harness is the
operating system, and a skill is software installed into that environment, then
task success only tells us whether the stack produced the right final output.
Skill coverage asks a more diagnostic question: when a task made some part of
the skill relevant, was that instruction actually exercised, and did the agent
follow it? In this sense, both code coverage and skill coverage measure test
adequacy beyond end-to-end pass rates.

The difference is that code coverage operates over hard structure, while skill
coverage operates over soft behavioral specifications. Lines, branches, and
functions are given by syntax; Skill Behavior Constraints (\sbcs) must be
extracted from natural-language skill prose and normalized into observable
conditional requirements. A line of code is covered when it executes, but an
\sbc is covered only when its applicability condition is instantiated in the
trajectory. Once triggered, the behavior may pass or fail; both cases are
covered, because both mean the benchmark tested that instruction. Only
\textsc{Not covered} means the behavior was never exercised. This makes skill
coverage harder to measure: the oracle is semantic, the evidence is scattered
across messages, tool calls, files, logs, and artifacts, and repeated agent runs
may trigger different parts of the same skill. Thus, skill coverage is not
line coverage for prompts, but coverage for agent-facing behavioral
specifications.

\subsection{Threats to Validity}

\subsubsection{Judge and Trace Validity}
Coverage labeling depends on semantic judgment over observable trajectories. The
LLM judge may miss evidence, infer compliance from weak traces, or misclassify
whether an applicability condition was instantiated. We reduce this risk by
checking human--human agreement, validating the DeepSeek V4 Flash judge against
human labels, and then freezing the judging protocol before applying it at
scale. The labels still reflect a validated but fallible protocol. They also use
only evidence visible to an external evaluator, such as messages, tool calls,
command outputs, file states, generated artifacts, and final responses. This
choice makes the measurement auditable, but it can understate behavior that
occurred without visible evidence.

\subsubsection{Conclusion Validity}
The main conclusion-validity threat for RQ3 is that a recovered task outcome
could be caused by stochastic variation rather than by the coverage-guided
skill-emphasis edit. We mitigate this threat by rerunning each strengthened task
three times under the original benchmark verifier. The repeated attempts make
the reported recovery rate a stability signal rather than a single lucky
trajectory: improvements are counted only through the same task verifier, and
the aggregate RQ3 result reflects behavior observed across repeated
strengthened executions.

\subsubsection{External Validity}
The empirical results come from the current SkillsBench coverage slice: 87
source tasks, 202 task-associated skill artifacts, five agent--model
configurations, and three reruns per configuration. The selected rows cover
multiple success regimes and agent harnesses, but they do not represent all
models, agents, skill libraries, or task domains. Other benchmarks may pair
skills with tasks differently, record different trajectory evidence, or use
skills written in different styles. The strengthening results are narrower
still, because they use failed diagnostic tasks and constraints that survived
triage as plausible instruction-emphasis targets. The numerical coverage and
recovery rates should therefore be read as evidence about this setting. Broader
claims require additional benchmarks, independent skill collections, and
controlled strengthening studies.

\section{Open Questions}
\label{sec:future-work}
We see skill coverage as a first adequacy lens for reusable agent skills, one that
raises questions task-level evaluation alone does not answer. As RQ2 showed, task
outcomes do not reveal which documented behaviors were exercised. Making this
behavior observable, rather than hidden behind a task verdict, raises new questions
about how skills should be tested, reported, and maintained.

\emph{Test generation.} Uncovered constraints are natural generation targets:
varying an existing task or generating a new one to instantiate an uncovered
condition could turn skill evaluation into a coverage-driven test-generation loop,
approaching complete coverage of documented behavior rather than merely adding
tasks.

\emph{Skill test reports.} Beyond a single pass rate, coverage evidence could let
users inspect a skill before reusing it: which instructions are tested, which
branches remain untested, and which behaviors failed. Skills could then be trusted
by tested behavior rather than by download count.

\emph{Skill maintenance.} Failed constraints point to specific instructions that
could be repaired, by making them more salient, adding an ordering gate, or
attaching an evidence requirement. Our RQ3 probe is early evidence this helps:
re-emphasizing failed constraints recovered 16.0\% of failed tasks, suggesting some
failures stem from unused rather than missing skill knowledge. How to turn coverage
labels into a reliable repair workflow, and how to prioritize which constraints to
fix first, remain open.

We intend skill coverage as an initial step toward systematic testing of reusable agent skills, one we hope reframes skill quality as a question of tested behavior rather than task success alone.
\section{Related Work}

\subsection{Agent Skills and Skill Benchmarks}
Agent skills are emerging as reusable natural-language artifacts for extending
LLM agents with procedural knowledge, tool-use guidance, and workflow
expectations~\citep{anthropic2026agentskills,Ling2026AgentSA,Zhou2026ACS}.
Recent benchmarks and studies evaluate whether such skills improve downstream
task performance, how well agents use skills in realistic settings, and how
skills can be generated or accumulated over time~\citep{Li2026SkillsBenchBH,
Zhong2026SkillLearnBenchBC,Liu2026HowWD,Huang2025CASCADECA}. Other work audits
open skill ecosystems and highlights the growing importance of skill quality
as skills become shared across agents and tasks~\citep{Ying2026OpenSkillEvalAA}.
These efforts establish that skills are practically useful and worth
benchmarking. Our work addresses a complementary question: once a reusable
skill is evaluated through benchmark executions, which documented behaviors in
the skill have actually been exercised with judgeable evidence? Our
task-outcome-conditioned analysis further shows why this question is not
answered by task outcome alone: task success is related to coverage, but it does not
exhaust the documented skill denominator.

\subsection{Requirements, Traceability, and Test Adequacy}
Our formulation builds on the view that requirements should be verifiable,
traceable, and connected to desired system behavior~\citep{zave1997four,
gotel1994analysis,iso29148}. Because skill documents are informal and
heterogeneous, we use EARS-style controlled natural language as a lightweight
normalization layer for extracting source-grounded behavioral constraints from
skill prose~\citep{mavin2009ears}. The coverage side of our work follows the
testing tradition that adequacy criteria should describe what a test suite has
exercised, rather than only reporting individual test outcomes~\citep{
goodenough1975toward,weyuker1986axiomatizing,weyuker1988evaluation,
zhu1997coverage}. Requirements-based coverage further motivates reporting a
denominator over specified behavior rather than over incidental execution
artifacts~\citep{whalen2006coverage}. We adapt these ideas to agent skills by
treating the skill document as the artifact under test and agent trajectories
as the evidence surface. This framing separates verifier success from
source-grounded skill behavior coverage: one asks whether a task outcome was
achieved, while the other asks which documented behaviors received judgeable
trajectory evidence.

\subsection{Oracle-Aware Coverage and LLM Judges}
Skill coverage also relates to the oracle problem: an execution is useful for
testing only when it exposes evidence that can be checked by an oracle
\citep{barr2015oracle}. Checked coverage formalizes a similar idea for
programs by asking whether executed elements influence oracle-checked values
\citep{schuler2011assessing}. We reinterpret this principle for natural
language skill constraints: a documented behavior is covered only when the
trajectory instantiates the relevant condition and contains observable evidence
sufficient to judge the constrained behavior. At benchmark scale, this requires many fine-grained attribution
decisions. Prior LLM-as-a-judge work shows that automatic judges can be useful
when first validated against human or expert judgments~\citep{
Zheng2023JudgingLW,Liu2023GEvalNE,dubois2024lengthcontrolled}. Fine-grained
rubric-based evaluators further show how LLM judges can apply predefined
criteria at scale~\citep{Ye2023FLASKFL,ICLR2024_80348535,
Starace2025PaperBenchEA,Arora2025HealthBenchEL}. Our evaluation follows this
validation-before-scale pattern by using LLM-based constraint labels only after
checking both three-label agreement and agreement on the coverage denominator
against human labels.

\section{Conclusion}

Reusable agent skills are becoming shared procedural artifacts, but current
skill benchmarks primarily report whether agents complete tasks. This paper
shows that task success is not enough to assess whether a skill artifact has
been adequately tested. We introduced \textit{skill coverage}: a
trajectory-based adequacy metric that extracts source-grounded skill behavior
constraints from skill documents and asks which constraints become observable in
an agent execution. By separating \emph{not covered}, \emph{Pass}, and \emph{Fail}
labels, the metric distinguishes missing benchmark pressure from observed
skill-following success and observed skill-use breakdown.

Applying this framework to SkillsBench gives a more fine-grained account of
what existing evaluations exercise. After validating the coverage judge against
human labels, we measured five agent--model configurations over 87 source tasks
and three reruns per configuration. The benchmark trajectories covered only
38.66--45.51\% of the audited skill behavior denominator on average, even
though the same rows differed substantially in task success. This result shows
that leaderboard outcomes and artifact-level adequacy answer different
questions: a task run can be useful for measuring task completion while still
leaving most documented skill behaviors unobserved.

The labels are also useful beyond measurement. In failed tasks, \emph{Fail}
verdicts identify cases where a documented skill behavior became applicable but
the agent violated or omitted it. Using those failed constraints only to
emphasize original skill instructions produced a 16.0\% average recovery rate
over three strengthened-skill attempts, and selected strengthened constraints
achieved 43.8--66.2\% covered-pass rates when re-judged in the new
trajectories. These findings support a diagnostic, not causal, conclusion:
skill coverage can expose actionable instruction-following signals in selected
failed tasks. More broadly, coverage labels can guide both sides of skill
maintenance: uncovered constraints suggest where benchmark tasks should be
varied or generated, while failed constraints suggest where skill instructions
or agent harness behavior should be inspected before a skill is reused.

\section*{Data Availability}

The data and study materials needed to inspect and reproduce the empirical
results reported in this paper are available in the public artifact repository:
\url{https://github.com/shuaijiumei/skill-coverage-artifact}.

\balance
\bibliography{references}

@article{zave1997four,
  author  = {Zave, Pamela and Jackson, Michael},
  title   = {Four Dark Corners of Requirements Engineering},
  journal = {ACM Transactions on Software Engineering and Methodology},
  volume  = {6},
  number  = {1},
  pages   = {1--30},
  year    = {1997}
}

@inproceedings{gotel1994analysis,
  author    = {Gotel, Orlena C. Z. and Finkelstein, Anthony C. W.},
  title     = {An Analysis of the Requirements Traceability Problem},
  booktitle = {Proceedings of the IEEE International Conference on Requirements Engineering},
  pages     = {94--101},
  year      = {1994}
}

@misc{iso29148,
  author       = {{ISO/IEC/IEEE}},
  title        = {{ISO/IEC/IEEE 29148:2018}: Systems and Software Engineering---Life Cycle Processes---Requirements Engineering},
  organization = {ISO/IEC/IEEE},
  year         = {2018}
}

@inproceedings{mavin2009ears,
  author    = {Mavin, Alistair and Wilkinson, Philip and Harwood, Adrian and Novak, Mark},
  title     = {Easy Approach to Requirements Syntax {(EARS)}},
  booktitle = {Proceedings of the 17th IEEE International Requirements Engineering Conference},
  pages     = {317--322},
  year      = {2009}
}

@article{weyuker1988evaluation,
  author  = {Weyuker, Elaine J.},
  title   = {The Evaluation of Program-Based Software Test Data Adequacy Criteria},
  journal = {Communications of the ACM},
  volume  = {31},
  number  = {6},
  pages   = {668--675},
  year    = {1988}
}

@article{zhu1997coverage,
  author  = {Zhu, Hong and Hall, Patrick A. V. and May, John H. R.},
  title   = {Software Unit Test Coverage and Adequacy},
  journal = {ACM Computing Surveys},
  volume  = {29},
  number  = {4},
  pages   = {366--427},
  year    = {1997}
}

@article{goodenough1975toward,
  author  = {Goodenough, John B. and Gerhart, Susan L.},
  title   = {Toward a Theory of Test Data Selection},
  journal = {IEEE Transactions on Software Engineering},
  volume  = {SE-1},
  number  = {2},
  pages   = {156--173},
  year    = {1975},
  doi     = {10.1109/TSE.1975.6312836}
}

@article{weyuker1986axiomatizing,
  author  = {Weyuker, Elaine J.},
  title   = {Axiomatizing Software Test Data Adequacy},
  journal = {IEEE Transactions on Software Engineering},
  volume  = {SE-12},
  number  = {12},
  pages   = {1128--1138},
  year    = {1986},
  doi     = {10.1109/TSE.1986.6313008}
}

@inproceedings{schuler2011assessing,
  author    = {Schuler, David and Zeller, Andreas},
  title     = {Assessing Oracle Quality with Checked Coverage},
  booktitle = {Proceedings of the 4th International Conference on Software Testing, Verification and Validation},
  pages     = {90--99},
  year      = {2011},
  doi       = {10.1109/ICST.2011.32}
}

@inproceedings{whalen2006coverage,
  author    = {Whalen, Michael W. and Rajan, Ajitha and Heimdahl, Mats P. E. and Miller, Steven P.},
  title     = {Coverage Metrics for Requirements-Based Testing},
  booktitle = {Proceedings of the 2006 International Symposium on Software Testing and Analysis},
  pages     = {25--36},
  year      = {2006},
  doi       = {10.1145/1146238.1146242}
}

@article{barr2015oracle,
  author  = {Barr, Earl T. and Harman, Mark and McMinn, Phil and Shahbaz, Muzammil and Yoo, Shin},
  title   = {The Oracle Problem in Software Testing: A Survey},
  journal = {IEEE Transactions on Software Engineering},
  volume  = {41},
  number  = {5},
  pages   = {507--525},
  year    = {2015},
  doi     = {10.1109/TSE.2014.2372785}
}

@article{Li2026SkillsBenchBH,
  title={SkillsBench: Benchmarking How Well Agent Skills Work Across Diverse Tasks},
  author={Xiangyi Li and Wenbo Chen and Yimin Liu and Shenghan Zheng and Xiaokun Chen and Yifeng He and Yubo Li and B. You and Haotian Shen and Jiankai Sun and Shuyi Wang and Qunhong Zeng and Di Wang and Xuandong Zhao and Yuanli Wang and Roey Ben Chaim and Zonglin Di and Yi Gao and Junwei He and Yizhuo He and Liqiang Jing and Luyang Kong and Xin Lan and Jiachen Li and Songlin Li and Yijiang Li and Yue Lin and Xinyi Liu and Xuanqing Liu and Hao Lyu and Zexiong Ma and Bowei Wang and Runhui Wang and Tianyu Wang and Wengao Ye and Yue Zhang and Hanwen Xing and Y. Xue and Steven Dillmann and Han Lee},
  journal={ArXiv},
  year={2026},
  volume={abs/2602.12670},
  url={https://api.semanticscholar.org/CorpusID:285606595}
}

@article{Zhong2026SkillLearnBenchBC,
  title={SkillLearnBench: Benchmarking Continual Learning Methods for Agent Skill Generation on Real-World Tasks},
  author={Shan Zhong and Yiming Lu and Jingjie Ning and Yibing Wan and Lihang Feng and Yuyi Ao and Leonardo F. R. Ribeiro and Markus Dreyer and Sean Ammirati and Chenyan Xiong},
  journal={ArXiv},
  volume={abs/2604.20087},
  year={2026},
  url={https://api.semanticscholar.org/CorpusID:287669099}
}

@article{Ling2026AgentSA,
  author  = {G. Ling and Shan Zhong and R. Huang},
  title   = {Agent Skills: A Data-Driven Analysis of Claude Skills for Extending Large Language Model Functionality},
  journal = {ArXiv},
  volume  = {abs/2602.08004},
  year    = {2026}
}

@article{Zhou2026ACS,
  author  = {Yingli Zhou and Wang Shu and Yaodong Su and Wenchuan Du and Yixiang Fang and Xuemin Lin},
  title   = {A Comprehensive Survey on Agent Skills: Taxonomy, Techniques, and Applications},
  journal = {ArXiv},
  volume  = {abs/2605.07358},
  year    = {2026}
}

@article{Liu2026HowWD,
  author  = {Yujian Liu and Jiabao Ji and Li An and T. Jaakkola and Yang Zhang and Shiyu Chang},
  title   = {How Well Do Agentic Skills Work in the Wild: Benchmarking LLM Skill Usage in Realistic Settings},
  journal = {ArXiv},
  volume  = {abs/2604.04323},
  year    = {2026}
}

@Article{Huang2025CASCADECA,
 author = {Xu Huang and Junwu Chen and Yuxing Fei and Zhuohan Li and P. Schwaller and Gerbrand Ceder},
 booktitle = {arXiv.org},
 journal = {ArXiv},
 title = {CASCADE: Cumulative Agentic Skill Creation through Autonomous Development and Evolution},
 volume = {abs/2512.23880},
 year = {2025}
}

@article{Ying2026OpenSkillEvalAA,
  author  = {Jiahao Ying and Bo Ai and Wei Tang and Siyuan Liu and Yixin Cao},
  title   = {OpenSkillEval: Automatically Auditing the Open Skill Ecosystem for LLM Agents},
  journal = {ArXiv},
  volume  = {abs/2605.23657},
  year    = {2026}
}

@misc{anthropic2026agentskills,
  title        = {Agent Skills},
  author       = {{Anthropic}},
  year         = {2026},
  howpublished = {\url{https://docs.claude.com/en/docs/agents-and-tools/agent-skills}},
  note         = {Accessed: 2026-05-19}
}

@misc{openai2026codex,
  author       = {{OpenAI}},
  title        = {Codex},
  year         = {2026},
  howpublished = {\url{https://openai.com/codex/}},
  note         = {Accessed: 2026-06-30}
}

@misc{openai2026gpt55systemcard,
  author       = {{OpenAI}},
  title        = {{GPT-5.5} System Card},
  year         = {2026},
  howpublished = {\url{https://openai.com/index/gpt-5-5-system-card/}},
  note         = {Accessed: 2026-06-30}
}

@misc{anthropic2026claudecode,
  author       = {{Anthropic}},
  title        = {Claude Code Overview},
  year         = {2026},
  howpublished = {\url{https://docs.anthropic.com/en/docs/claude-code/overview}},
  note         = {Accessed: 2026-06-30}
}

@misc{anthropic2026claudeopus47systemcard,
  author       = {{Anthropic}},
  title        = {Claude Opus 4.7 System Card},
  year         = {2026},
  howpublished = {\url{https://www.anthropic.com/claude-opus-4-7-system-card}},
  note         = {Accessed: 2026-06-30}
}

@inproceedings{Wang2024OpenHandsAO,
  title     = {OpenHands: An Open Platform for {AI} Software Developers as Generalist Agents},
  author    = {Xingyao Wang and Boxuan Li and Yufan Song and Frank F. Xu and Xiangru Tang and Mingchen Zhuge and Jiayi Pan and Yueqi Song and Bowen Li and Jaskirat Singh and Hoang H. Tran and Fuqiang Li and Ren Ma and Mingzhang Zheng and Bill Qian and Yanjun Shao and Niklas Muennighoff and Yizhe Zhang and Binyuan Hui and Junyang Lin and Robert Brennan and Hao Peng and Heng Ji and Graham Neubig},
  booktitle = {The Thirteenth International Conference on Learning Representations},
  year      = {2025},
  url       = {https://api.semanticscholar.org/CorpusID:271404219}
}

@misc{deepseekai2026deepseekv4,
  title         = {{DeepSeek-V4}: Towards Highly Efficient Million-Token Context Intelligence},
  author        = {{DeepSeek-AI} and Anyi Xu and Bangcai Lin and others},
  year          = {2026},
  eprint        = {2606.19348},
  archivePrefix = {arXiv},
  primaryClass  = {cs.CL},
  url           = {https://arxiv.org/abs/2606.19348}
}

@article{glm5team2026glm5,
  title   = {{GLM-5}: from Vibe Coding to Agentic Engineering},
  author  = {{GLM-5 Team} and Aohan Zeng and Xin Lv and others},
  journal = {ArXiv},
  volume  = {abs/2602.15763},
  year    = {2026},
  url     = {https://api.semanticscholar.org/CorpusID:285659808}
}

@misc{zai2026glm51,
  author       = {{Z.ai}},
  title        = {{GLM-5.1}},
  year         = {2026},
  howpublished = {\url{https://docs.z.ai/guides/llm/glm-5.1}},
  note         = {Accessed: 2026-06-30}
}

@Article{Zheng2023JudgingLW,
 author = {Lianmin Zheng and Wei-Lin Chiang and Ying Sheng and Siyuan Zhuang and Zhanghao Wu and Yonghao Zhuang and Zi Lin and Zhuohan Li and Dacheng Li and E. Xing and Haotong Zhang and Joseph E. Gonzalez and Ion Stoica},
 booktitle = {Neural Information Processing Systems},
 journal = {ArXiv},
 title = {Judging LLM-as-a-judge with MT-Bench and Chatbot Arena},
 volume = {abs/2306.05685},
 year = {2023}
}

@Article{Liu2023GEvalNE,
 author = {Yang Liu and Dan Iter and Yichong Xu and Shuo Wang and Ruochen Xu and Chenguang Zhu},
 booktitle = {Conference on Empirical Methods in Natural Language Processing},
 journal = {ArXiv},
 title = {G-Eval: NLG Evaluation using GPT-4 with Better Human Alignment},
 volume = {abs/2303.16634},
 year = {2023}
}

@Article{Ye2023FLASKFL,
 author = {Seonghyeon Ye and Doyoung Kim and Sungdong Kim and Hyeonbin Hwang and Seungone Kim and Yongrae Jo and James Thorne and Juho Kim and Minjoon Seo},
 booktitle = {International Conference on Learning Representations},
 journal = {ArXiv},
 title = {FLASK: Fine-grained Language Model Evaluation based on Alignment Skill Sets},
 volume = {abs/2307.10928},
 year = {2023}
}

@misc{dubois2024lengthcontrolled,
      title={Length-Controlled AlpacaEval: A Simple Way to Debias Automatic Evaluators},
      author={Yann Dubois and Bal{\'a}zs Galambosi and Percy Liang and Tatsunori B. Hashimoto},
      year={2024},
      eprint={2404.04475},
      archivePrefix={arXiv},
      primaryClass={cs.LG},
      url={https://arxiv.org/abs/2404.04475},
}

@inproceedings{ICLR2024_80348535,
 author = {Kim, Seungone and Shin, Jay and cho, yejin and Jang, Joel and Longpre, Shayne and Lee, Hwaran and Yun, Sangdoo and Shin, Ryan, S and Kim, Sungdong and Thorne, James and Seo, Minjoon},
 booktitle = {International Conference on Learning Representations},
 editor = {B. Kim and Y. Yue and S. Chaudhuri and K. Fragkiadaki and M. Khan and Y. Sun},
 pages = {29927--29962},
 title = {Prometheus: Inducing Fine-Grained Evaluation Capability in Language Models},
 url = {https://proceedings.iclr.cc/paper_files/paper/2024/file/803485352e61e3ebf41221e4776c9fd4-Paper-Conference.pdf},
 volume = {2024},
 year = {2024}
}

@Article{Starace2025PaperBenchEA,
 author = {Giulio Starace and Oliver Jaffe and Dane Sherburn and J. Aung and Jun Shern Chan and Leon Maksin and Rachel Dias and E. Mays and Benjamin Kinsella and Wyatt Thompson and Johannes Heidecke and Amelia Glaese and Tejal Patwardhan},
 booktitle = {International Conference on Machine Learning},
 journal = {ArXiv},
 title = {PaperBench: Evaluating AI's Ability to Replicate AI Research},
 volume = {abs/2504.01848},
 year = {2025}
}

@Article{Arora2025HealthBenchEL,
 author = {Rahul K. Arora and Jason Wei and Rebecca Soskin Hicks and Preston Bowman and J. Q. Candela and Foivos Tsimpourlas and Michael Sharman and Meghan Shah and Andrea Vallone and Alex Beutel and Johannes Heidecke and Karan Singhal},
 booktitle = {arXiv.org},
 journal = {ArXiv},
 title = {HealthBench: Evaluating Large Language Models Towards Improved Human Health},
 volume = {abs/2505.08775},
 year = {2025}
}

@article{krejcie1970determining,
  author  = {Robert V. Krejcie and Daryle W. Morgan},
  title   = {Determining Sample Size for Research Activities},
  journal = {Educational and Psychological Measurement},
  volume  = {30},
  number  = {3},
  pages   = {607--610},
  year    = {1970},
  doi     = {10.1177/001316447003000308}
}

@misc{clawhub2026topskills,
  author       = {{ClawHub}},
  title        = {ClawHub Top Skills Listing},
  year         = {2026},
  howpublished = {\url{https://clawhub.ai/}},
  note         = {Accessed: 2026-06-28}
}

@misc{clawhub2026gog,
  author       = {{ClawHub}},
  title        = {Gog Skill},
  year         = {2026},
  howpublished = {\url{https://clawhub.ai/steipete/skills/gog}},
  note         = {Accessed: 2026-06-28}
}

@misc{clawhub2026nanopdf,
  author       = {{ClawHub}},
  title        = {Nano PDF Skill},
  year         = {2026},
  howpublished = {\url{https://clawhub.ai/steipete/skills/nano-pdf}},
  note         = {Accessed: 2026-06-28}
}

@article{Sumers2023CognitiveAF,
  title={Cognitive Architectures for Language Agents},
  author={Theodore R. Sumers and Shunyu Yao and Karthik Narasimhan and Thomas L. Griffiths},
  journal={Trans. Mach. Learn. Res.},
  year={2023},
  volume={2024},
  url={https://api.semanticscholar.org/CorpusID:261556862}
}

@article{Sutton1999BetweenMA,
  title={Between MDPs and Semi-MDPs: A Framework for Temporal Abstraction in Reinforcement Learning},
  author={Richard S. Sutton and Doina Precup and Satinder Singh},
  journal={Artif. Intell.},
  year={1999},
  volume={112},
  pages={181-211},
  url={https://api.semanticscholar.org/CorpusID:76564}
}

@inproceedings{Gao2004TestingComponentBasedSoftware,
  title={Testing Component-Based Software -- Issues, Challenges, and Solutions},
  author={Jerry Zeyu Gao and Ye Wu},
  booktitle={COTS-Based Software Systems},
  year={2004},
  pages={2--2},
  doi={10.1007/978-3-540-24645-9_2},
  url={https://api.semanticscholar.org/CorpusID:273339531}
}
\bibliographystyle{IEEEtran}

\end{document}